\documentclass{article} 
\usepackage{iclr2025_conference,times}
\usepackage{xcolor}
\usepackage{graphicx}
\usepackage{amsthm}
\usepackage{algorithm}
\usepackage{booktabs}
\usepackage{multirow}
\usepackage{algpseudocode}
\usepackage{subfigure} 

\usepackage{pdflscape} 
\usepackage{longtable} 

\usepackage{amsmath,amsfonts,bm}

\def\eqref#1{equation~\ref{#1}}

\def\1{\bm{1}}

\DeclareMathAlphabet{\mathsfit}{\encodingdefault}{\sfdefault}{m}{sl}
\SetMathAlphabet{\mathsfit}{bold}{\encodingdefault}{\sfdefault}{bx}{n}

\usepackage{hyperref}
\usepackage{url}

\newtheorem{definition}{Definition}

\title{Agent models: Internalizing Chain-of-Action Generation into Reasoning models}
\author{Yuxiang Zhang\textsuperscript{†}, Yuqi Yang\textsuperscript{†}, Jiangming Shu, Xinyan Wen \& Jitao Sang\thanks{Corresponding author.} \\
School of Computer Science and Technology\\
Beijing Jiaotong University\\
Beijing, China\\
\texttt{\{yuxiangzhang, yqyang, jiangmingshu, xinyanwen, jtsang\}@bjtu.edu.cn} \\
\small
}

\iclrfinalcopy 
\begin{document}

\maketitle
\footnotetext[1]{† Equal contribution.}

\begin{abstract}
Traditional agentic workflows rely on external prompts to manage interactions with tools and the environment, which limits the autonomy of reasoning models. We position \emph{Large Agent Models (LAMs)} that internalize the generation of \emph{Chain-of-Action (CoA)}, enabling the model to autonomously decide when and how to use external tools. Our proposed AutoCoA framework combines supervised fine-tuning (SFT) and reinforcement learning (RL), allowing the model to seamlessly switch between reasoning and action while efficiently managing environment interactions. Main components include step-level action triggering, trajectory-level CoA optimization, and an internal world model to reduce real-environment interaction costs. Evaluations on open-domain QA tasks demonstrate that AutoCoA-trained agent models significantly outperform ReAct-based workflows in task completion, especially in tasks that require long-term reasoning and multi-step actions. Code and dataset are available at
\url{https://github.com/ADaM-BJTU/AutoCoA}.
\end{abstract}

\section{Introduction}

OpenAI has outlined five progressive stages on the path to Artificial General Intelligence (AGI). The first stage, characterized as \textbf{Chatbot}, is exemplified by \textcolor{blue}{Large Language Models (LLMs)} like \emph{GPT-3.5} and \emph{GPT-4}~\cite{openai_gpt4_2023}. The second stage, termed \textbf{Reasoner}, introduces \textcolor{blue}{Large Reasoning Models (LRMs)} such as \emph{o1}~\cite{openai_o1_system_card} and \emph{o3}.
Recently, OpenAI released \emph{Operator}~\cite{openai_operator_2025} and \emph{Deep Research}~\cite{openai_ozdr_2025}, signaling the arrival of the third stage: \textbf{Agent}. These systems reportedly combine reasoning with autonomous tool usage, enabling independent execution of multi-round workflows by interacting with the real-world environment.

It is believed that the technology behind \emph{Operator} and \emph{Deep Research} is not merely integrating existing LLMs or LRMs with agentic workflows (e.g., ReAct~\cite{yao2022react}, Reflexion~\cite{shinn2023reflexion}). Instead, it represents a further upgrade in model capabilities: the new models are capable of long-term planning, tool manipulation, and environmental interaction. Taking \emph{Deep Research} as an example, most currently available open-source versions, such as \emph{Open Deep Research}~\cite{huggingfaceopendeepresearch} by Hugging Face, primarily rely on agentic workflows to orchestrate task planning and tool usage. The switching between thought and action is triggered by preset workflows, making it a ``passive'' behavior. In contrast, OpenAI's Deep Research enhances the multi-turn tool usage capability of reasoning models. The switching between thought and action is based on the model's inherent behaviors, where the model ``actively'' decides when and how to take action~\footnote{~In an agentic workflow employing reasoning models, the agent also triggers actions based on its own behavior during reasoning, but this switching is only based on the preceding context. In contrast, the model behind OpenAI's Deep Research has undergone end-to-end fine-tuning on the basis of reasoning models for tool usage capabilities. It considers not only the preceding context but also the subsequent context, explicitly trained with task completion as the optimization objective.}. This allows for a longer chain of thought-action loops, enabling the completion of more complex knowledge search tasks. We believe this type of \textbf{tool-augmented reasoning model} deserves a distinct name, and we propose calling it \textcolor{blue}{Large Agent Models (LAMs)}.

Task planning and tool usage are two core capabilities of an Agent. The reasoning model internalizes the capability of task planning, i.e., generating Chain-of-Thought (CoT). By further enhancing and internalizing the capability of tool usage, i.e., generating \emph{Chain-of-Action (CoA)}, the reasoning model evolves into an \textbf{Agent model}. 

\begin{figure}[t] 
  \centering 
  \includegraphics[width=0.96\textwidth]{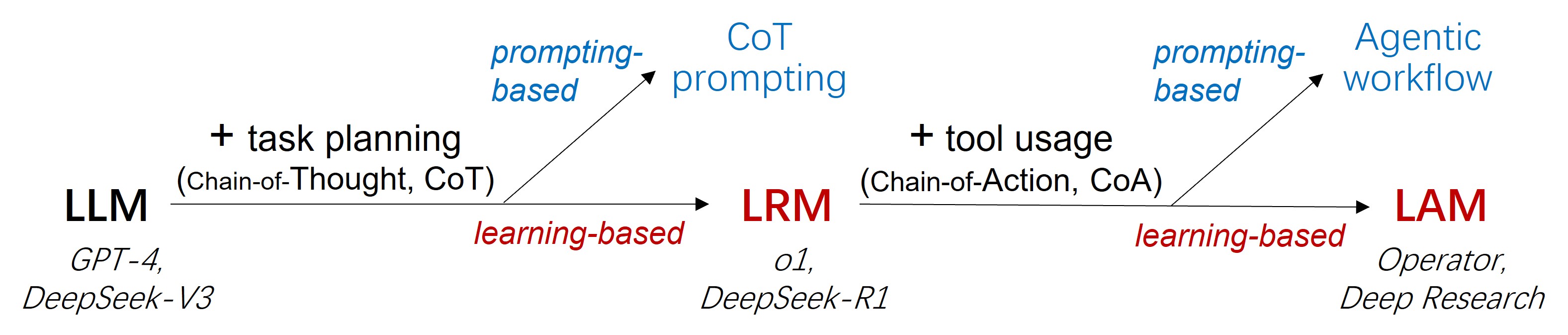} 
  \caption{The path from Large Language Model (LLM) to Large Reasoning Model (LRM) and the forthcoming Large Agent Model (LAM).} 
  \label{fig:1} 
\end{figure}

It is helpful to review the development of reasoning models to understand the relationship between agentic workflow and agent model. As illustrated in Fig.~\ref{fig:1}, a reasoner needs to add the ability to generate Chain-of-Thought (CoT) on top of the System-1 LLM. One approach is through prompting, leveraging in-context learning to ``force'' the model to output multi-step reasoning, commonly referred as \emph{CoT prompting}~\cite{cot_prompting_2022}. Another approach is learning-based~\cite{zhang2024o1codero1replicationcoding,zeng2024scaling}, whether through SFT alone, a combination of SFT followed by RL, or direct RL. This results in reasoning models like o1 and DeepSeek-R1. Unlike CoT prompting relying on external guidance to elicit multi-step reasoning, reasoning models learn the logics between thinking steps and internalize the CoT generation capability as a ``active'' model behavior.

Similarly, agents need to move beyond CoT reasoning to CoA generation, a paradigm dynamically interleaving thought and action. 
While \emph{agentic workflow} relies on prompting-based methods to simulate this interplay, agent model learns the coupled CoT and CoA patterns into its behavior, resulting in stronger logical coherence and seamless transitions between thought and action. The model initiates tool usage only when necessary and makes context-aware decisions on how to execute it. Harnessing tools to embody in digital/physical environments, agent models effectively bridge the gap between reasoning and action.

In fact, this paradigm shift from prompting to learning aligns with AI’s decades-long evolution from handcrafted rules to data-driven learning: from expert systems to classical machine learning and then to deep learning. Patterns learned from data exhibit better generalization and greater flexibility. We have already witnessed that reasoning models like OpenAI o1 and DeepSeek R1 have marginalized the role of explicit prompting, as they can better understand user intent and proactively make plans. Similarly, the shift from agentic workflows to agent models, which internalize CoA generation, is expected to gradually replace the static and scripted workflows. 

In this report, we introduce the recent attempt to enhance the multi-turn tool usage capability of reasoning models, that is, to interleave CoA generation within the CoT trajectories. Our preliminary study focuses on small-scale reasoning models, using search action as a testbed for open-domain question answering (QA) tasks. Initial results demonstrate that internalizing CoA generation substantially improves task completion rates compared to agentic workflows. Future work will involve experimental validation on larger models, more tool types, and a wider range of open-ended tasks.

\section{Agent model}
\subsection{Definition and Formalization}

\begin{figure}[t] 
  \centering 
  \includegraphics[width=0.8\textwidth]{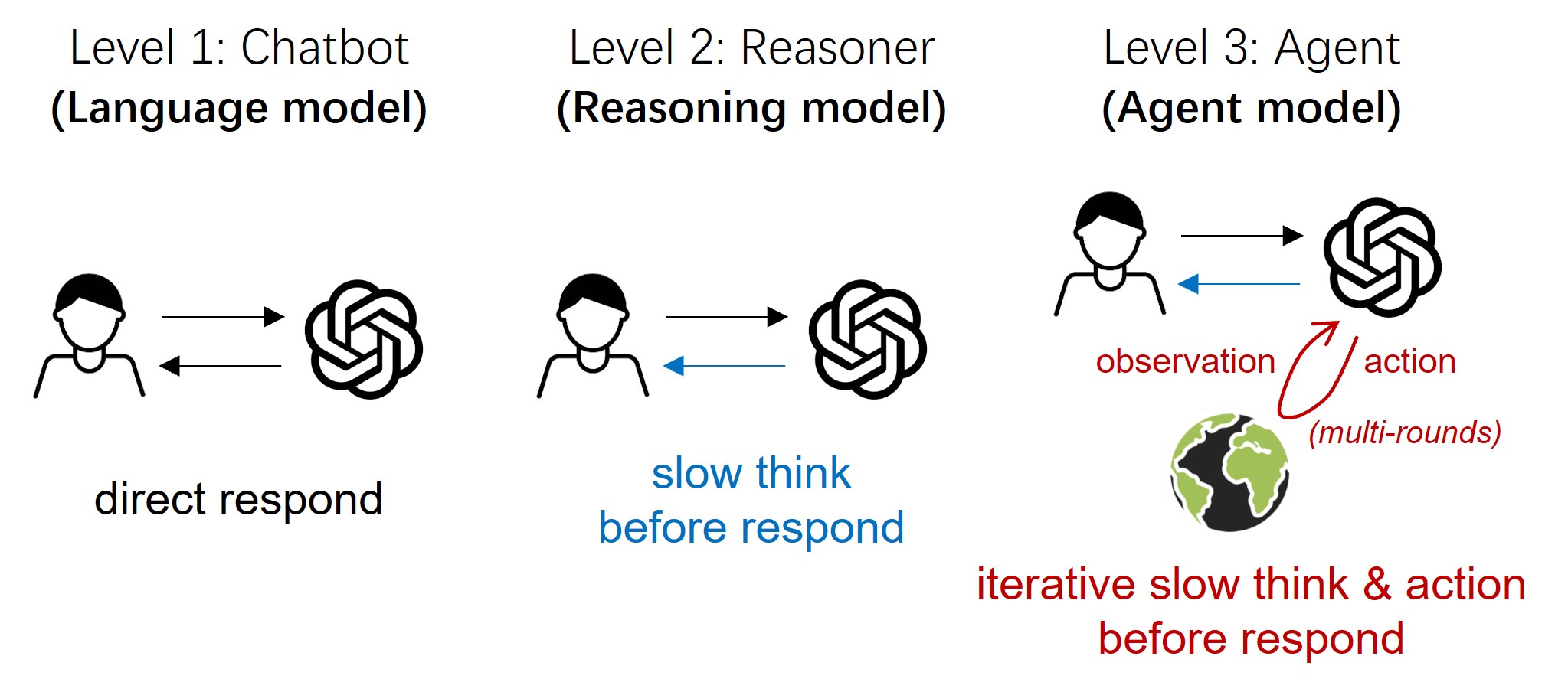} 
  \caption{Interaction paradigm: Chatbot vs. Reasoner vs. Agent.} 
  \label{fig:2} 
\end{figure}

Agent models are enhanced with the ability to use tools. This requires it to extend beyond its own reasoning behaviors and interact with the external environment. Fig.~\ref{fig:2} compares the interaction paradigms from chatbots to agent models. Both Chatbot and Reasoner focus exclusively on binary interactions between human and model. The difference lies in that reasoner-layer models engage in slow, deliberate reasoning before responding. While, the third layer, Agent, requires the model to think and act simultaneously, which takes the external environment into consideration. This creates a ternary structure of human-model-environment. Action, in this case, refers to interacting with the environment through tool usage to obtain feedback. After multiple rounds of thought, action, and observation, the final response is generated.

This applies to both Operator and Deep Research: Operator’s environment consists of operating systems and applications, and Deep Research currently operates within the environment of web-based information and data resources. 

We formally define \emph{Agent Model} as follows:\vspace{-1mm}
\begin{definition}
An \textbf{Agent Model} is a generative model built upon a reasoning model, enhanced through end-to-end task-oriented tool-augmented training. It produces sequences of interleaved reasoning (Chain-of-Thought) and action (Chain-of-Action) steps, where each action invokes a tool~\footnote{~Tools can be broadly categorized into two types: those for accessing external resources, such as real-time information and specialized databases via web browsers and API interfaces, and those for professional processing, such as complex computation and data analysis softwares.} to interact with the external environment. Observations from these interactions guide subsequent reasoning and actions until the task is completed.
\end{definition}

The inference of agent model can be formalized as a partially observable Markov decision process (POMDP), where the state is defined as \( [\mathbf{s}_0, \mathbf{tc}, \mathbf{x}_{1:n}] \), comprising:
(1) \( \mathbf{s}_0 \): the initial environment state (e.g., the state of a digital or physical system),
(2) \( \mathbf{tc} \): the task context (e.g., a user prompt),
(3) \( \mathbf{x}_{1:n} \): the sequence generated so far, combining tokens emitted by the model and responses from the environment.
Actions in this POMDP are the generation of the next token \( x_{n+1} \sim \pi_\theta(\cdot \mid \mathbf{s}_0, \mathbf{tc}, x_{1:n}) \), where \( \pi_\theta \) is the policy parameterized by \( \theta \).

At each step, the policy \( \pi_\theta \) can generate one of the following types of tokens:
\begin{itemize}
    \item \textcolor{blue}{\( \langle \text{think} \rangle \)}: Initiates or continues a reasoning step, contributing to the Chain-of-Thought (CoT), which represents internal deliberation or planning. Following this token, a reasoning sequence \( \mathbf{r} = [r_1, r_2, \dots, r_k] \) is generated, where each \( r_i \) is a token in the reasoning process.  
    \item \textcolor{blue}{\( \langle \text{action} \rangle \)}: Invokes a tool to interact with the environment, contributing to the Chain-of-Action (CoA). This token signals the transition from thought to action, deciding \textit{when to action}. Following this token, the model internally generates or acquires from environment:
    \begin{itemize}
        \item \textit{Action type}: specifies the tool from tool library  \(\tau \in \mathcal{T}\) (e.g., "search"),
        \item \textit{Action parameters}: \(\mathbf{p}\), defines how to execute the tool, deciding \textit{how to action} (e.g., detailed search query),
        \item \textit{Response by action invocation}: fetches the environment's response \( \mathbf{o} = \tau(\mathbf{p}) \), analogous to observations in ReAct. Note that this response is not generated by policy \( \pi_\theta \) but returned by interacting with the environment. \( \mathbf{o}\) will appended to the trajectory, providing context for subsequent turns. 
    \end{itemize}   
    \item \textcolor{blue}{\( \langle \text{answer} \rangle \)}: Outputs the final response, indicating task completion or termination due to exceeding a predefined budget. Following this token, an answer sequence \( \mathbf{a} = [a_1, a_2, \dots, a_m] \) is generated, representing the task solution.
\end{itemize}

The inference pipeline of agent model is shown in Algorithm \ref{alg:agent_model_pipeline}, 
outlining the step-by-step process by which the agent model generates its trajectory. 
The algorithm iteratively generates tokens based on the current state, deciding whether 
to engage in further reasoning, invoke a tool, or conclude with an answer. This pipeline 
ensures that the model alternates between internal reasoning (CoT) and external actions 
(CoA), allowing it to dynamically adapt its strategy based on environmental feedback. 
The loop continues until the task is completed or a predefined limit on reasoning-action 
iterations/contextual length is reached.

We can see that, unlike agentic workflows that rely on external prompts or scripts to manage interactions, the agent model internalizes the logic of tool usage as part of its learned behavior, enabling autonomous and context-aware decision-making for tool invocation.

\begin{algorithm}
\caption{Agent Model Inference Pipeline}
\label{alg:agent_model_pipeline}
\begin{algorithmic}[1]
\Require 
\Statex Policy model $\pi_\theta$, tool library $\mathcal{T}$, 
\Statex Task context $\mathbf{tc}$, max action steps $T$, max contextual length $N$ 
\Ensure 
\Statex Final answer $\mathbf{a}$
\vspace{2mm}
\State Initialize: $t \gets 0$, $n \gets \#(\mathbf{tc})$, $x_{1:n} \gets \mathbf{tc}$,  
\vspace{0.8mm}
\While{$t < T$ \textbf{and} $n < N$ \textbf{and} $\langle\text{answer}\rangle$ not in $x_{1:n}$}
    \State Generate token: $x_{n+1} \sim \pi_\theta(\cdot \mid x_{1:n})$
    \State $x_{1:n+1} \gets x_{1:n} \circ x_{n+1}$, $n \gets n+1$ 
    
    \If{$x_{n} = \textcolor{blue}{\langle\text{think}\rangle}$} \Comment{\textcolor{blue}{Generate CoT}}
        \While{next token $\notin \{\langle\text{action}\rangle, \langle\text{answer}\rangle\}$}
            \State $r \sim \pi_\theta(\cdot \mid x_{1:n})$ 
            \State $x_{1:n+1} \gets x_{1:n} \circ r$, $n \gets n+1$ 
        \EndWhile
        
    \ElsIf{$x_{n} = \textcolor{blue}{\langle\text{action}\rangle}$} \Comment{\textcolor{blue}{Generate CoA}}
        \State $t \gets t + 1$ 
        \State Generate $\mathbf{c} \sim \pi_\theta(\cdot \mid x_{1:n})$, $k=\#(\mathbf{c})$ \Comment{Generate action type}
        \State $x_{1:n+k} \gets x_{1:n} \circ \tau$, $n \gets n+k$
        \State Select tool: $\tau \gets \mathcal{T}[\mathbf{c}]$ \Comment{\textit{map action type tokens to specific tool}}
        \State Generate $\mathbf{p} \sim \pi_\theta(\cdot \mid x_{1:n})$, $k=\#(\mathbf{p})$ \Comment{Generate action parameter}
        \State $x_{1:n+k} \gets x_{1:n} \circ \mathbf{p}$, $n \gets n+k$
                \State Execute: $\mathbf{o} \gets \tau(\mathbf{p})$, $k=\#(\mathbf{p})$ \Comment{Invoke tool and acquire response}
        \State $x_{1:n+k} \gets x_{1:n} \circ  \mathbf{o}$, $n \gets n+k$
    \ElsIf{$x_{n} = \textcolor{blue}{\langle\text{answer}\rangle}$}
        \State Generate $\mathbf{a} \sim \pi_\theta(\cdot \mid x_{1:n})$, $n \gets n+1$ \Comment{\textcolor{blue}{Generate answer sequence}}
        \State \Return $\mathbf{a}$
    \EndIf
    
\EndWhile

\If{$\langle\text{answer}\rangle \notin x_{1:n}$} \Comment{Force termination and answer}
    \State $x_{1:n} \gets x_{1:n} \circ \langle\text{answer}\rangle$
    \State Generate $\mathbf{a} \sim \pi_\theta(\cdot \mid x_{1:n})$
\EndIf
\State \Return $\mathbf{a}$
\end{algorithmic}
\end{algorithm}

\subsection{Challenges}
The development of agent model requires end-to-end training on top of reasoning models to enhance its capabilities in tool usage and interaction with the external environment. This presents two fundamental challenges.

\textbf{Challenge 1: Balancing Reasoning and Action}. 
Reasoning models excel at generating Chains-of-Thought (CoT). Fine-tuning these models to enhance action execution capabilities poses a common problem: the risk of forgetting the model's CoT-generation ability~\cite{zhang2024openrft}. More critically, the key to interleaving CoT with Chain-of-Action (CoA) lies in teaching the model to trigger actions at appropriate times within the reasoning process. This ensures that reasoning and action complement each other effectively to jointly accomplish the task.

\textbf{Challenge 2: Managing External Environment Interaction}.
Interacting with the external environment through tool invocations may result in high training costs, low efficiency and even risks. Additionally, the dynamic nature of the external environment complicates the training process. For instance, during online reinforcement learning (RL), training delays and dynamic environmental changes will lead to instability in training, as the model needs to adapt to the changing environment.

OpenAI’s Deep Research exemplifies the practical challenges in agent model development. As officially claimed~\cite{openai_ozdr_2025}, Deep Research’s agent model is built by end-to-end reinforcement fine-tuning atop the flagship reasoning model \textit{o3}. The main action here is web search, and the environment consists of publicly available information on the internet.

Regarding Challenge 1, the model needs to autonomously determine during the reasoning process when its internal knowledge is insufficient, and when it needs to obtain information from the external environment. This requires the model to have a good understanding of its knowledge limitations and the ability to trigger actions at the right time.

As for Challenge 2, the cost of calling search engines and the dynamic nature of the network environment pose significant difficulties. The time and resources required for search calls can lead to high training costs and low efficiency. Moreover, the dynamic nature of the network environment makes it challenging for the model to adapt and respond in a timely manner.

In the next Section, we will use an open-domain QA task similar to Deep Research as the testbed, with the tool restricted to web search, to explore possible solutions. 

\section{Learn to Generate Chain-of-Action }

\subsection{Overview}
We introduce an agent model training solution, Automatic generation of Chain-of-Action (\textbf{AutoCoA}), which consists of two stages: Supervised Fine-Tuning (SFT) and Reinforcement Learning (RL), to internalize the Chain-of-Action (CoA) generation capability into reasoning models.

To address the challenge of reasoning-action balance, we mix pure CoT data (w/o tool usage) in both SFT and RL stages. This ensures that the model does not forget its reasoning capabilities and can adaptively decide whether to use action on tasks of different difficulties. Meanwhile, at the beginning of SFT, we add additional substage of injecting the tool-usage ability at step level, referred to as \emph{CoT+A}. Specifically, we generate contrastive data pairs by modifying a reasoning prefix in erroneous samples: one variant enables tool usage while the other does not. Successful task completion with tool invocation serves as a positive example, whereas failure without action forms a negative counterpart. Contrastive loss is then used to train the model on when to trigger actions.

To address the challenge of managing environment interaction, we explore the possibility of enabling the agent model to simulate the environment by incorporating an internal world model. Specifically, at the end of SFT, we added a sub-stage to train the policy model to simulate tool calls and generate the corresponding observations. During the RL stage, the model first generates its own observations based on the policy model and explores potential trajectories through extensive rollouts for fast convergence.  Then, only a limited number of tool calls are made to interact with the real environment, ensuring adaptation to real-world conditions.

The overall AutoCoA framework consists of the following stages:

\textbf{SFT Stage}
\begin{itemize}
    \item \textbf{CoT + A}: focusing on step-level action triggered by interleaving action tokens \textit{\( \langle \text{action} \rangle \)} within CoT sequences at critical reasoning junctures. Using contrastive loss, trajectories with necessary actions (e.g., invoking a search to resolve knowledge gaps) are compared against those without, teaching the model to autonomously decide \emph{when to action}.
        \item \textbf{CoT + CoA (w/ observation mask)}: the main training phase at the trajectory-level. During loss calculation for the policy model, the environment response is excluded. This teaches the model \emph{how to action}, including determining the action type and parameters.
        \item \textbf{CoT + CoA (w/o observation mask)}: training the model to predict environment responses (e.g., simulated search results) alongside actions. Joint optimization of action generation and observation prediction equips the policy with an implicit world model. This preconditions the model for the subsequent RL stage.
\end{itemize}

\textbf{RL Stage}
\begin{itemize}
    \item \textbf{CoT + CoA (simulated environment)}: after rolling out an \textit{\( \langle \text{action} \rangle \)}, the policy model itself continues to generate the observation leveraging the learned internal world model from SFT. This allows for extensive sampling to explore diverse action sequences, and rapid converging to a feasible solution space.
        \item \textbf{CoT + CoA (real environment)}: leveraging the improved exploration capabilities from the previous sub-stage, the model interacts with the real environment to receive observations. This substage helps the model adapt to the real-world environment and ensures robust deployment.
\end{itemize}

The AutoCoA framework enables progressive capability injection—from step-level action triggering to trajectory-level CoA optimizing, and from simulated exploration to real environment interaction. In the following subsections, we introduce in detail the implementation of each sub-stage.

\subsection{Supervised Fine-Tuning}

\subsubsection{Data Synthesis}

In this section, we use the training set of HotpotQA~\cite{yang2018hotpotqa} as our base dataset. HotpotQA is a multi-hop question-answering dataset containing questions and answers that require multi-step reasoning. We randomly selected 20,000 samples to construct our dataset \(\mathcal{D}\). Formally, each data point \(d \in \mathcal{D}\) contains the following elements:

\begin{itemize}
  \item Question \(q\): a natural language question requiring multi-step reasoning;
  \item Answer \(a\): the correct answer to the question.
\end{itemize}

To support the training requirements of the AutoCoA framework in the Supervised Fine-Tuning (SFT) stage, we designed three data synthesis methods to generate CoT data, CoA data, and CoT+A data, respectively. We used the DeepSeek-R1-Distill-Qwen-32B~\cite{guo2025deepseek} model for data generation. Regarding the environment for tool invocation, we set up a local wiki search engine based on FlashRAG~\cite{jin2024flashrag}. The integration of this tool enables the model to simulate the process of obtaining additional information through external resources during data generation, thereby training the model to intelligently trigger external tool calls when facing complex reasoning tasks, further enhancing its performance.

\textbf{CoT data}: We randomly selected 10,000 samples from dataset $\mathcal{D}$ as the base data. Using the DeepSeek-R1-Distill-Qwen-32B model with a CoT prompt designed to elicit step-by-step reasoning, we generated pure chain-of-thought processes without permitting external tool usage. The resulting CoT data captures the model’s sequential reasoning, encompassing problem decomposition, analysis of known information, and derivation of the final answer, all independent of external tools. These data are designed to maintain the model's basic reasoning abilities, ensuring it can handle questions that do not require external knowledge supplementation. Formally, we define the CoT data set as $\mathcal{D}_{CoT} = \{{(q, r_{CoT}, a) | (q, a) \in \mathcal{D}}\}$, where $r_{CoT}$ denotes the sequence of reasoning steps generated by the model.

\textbf{CoA data}: We selected another 10,000 samples from dataset $\mathcal{D}$ for construction. Using a CoA prompt that instructs the model to invoke search tools when encountering knowledge gaps, we enabled tool-assisted reasoning. The model triggers the ${\langle\text{action}\rangle}$ token to call a search tool, retrieves necessary information, and resumes reasoning based on the results. The generated CoA data captures the full chain: the question $q$, the reasoning process $r_{CoA}$ (including the action trigger ${\langle \text{action} \rangle}$, observation $o$ from the environment, and subsequent reasoning), and the final answer $a$. This data structure enables the model to learn when and how to trigger actions, as well as how to continue reasoning based on the returned observations. Formally, we define the set of CoA data as $\mathcal{D}_{CoA} = \{(q, r_{CoA}, a) | (q, a) \in \mathcal{D}\}$, where $r_{CoA}$ represents the interleaved sequence of reasoning, action, observation, and further reasoning.

\textbf{CoT+A data}: The CoT+A sample construction comprises two pairs of positive and negative examples. From the CoT and CoA data, we identified samples where either method yielded incorrect answers. For these, we generated contrastive pairs by altering the reasoning trajectory at a specific point: either inserting an action trigger ${\langle \text{action} \rangle}$ to enable tool usage or removing it to enforce internal reasoning. Each pair shares the same initial prefix but diverges on whether tool usage is permitted thereafter, leading to scenarios where either \emph{CoT is incorrect but CoA is correct} or \emph{CoA is incorrect but CoT is correct}. The former scenario highlights cases in which the model must leverage external tools to arrive at the correct solution, whereas the latter shows that indiscriminate tool usage can result in errors. We train these contrastive pairs using contrastive loss functions to help the model learn when to trigger actions and when to rely solely on its internal reasoning.

For the training process, we initially selected 1,500 CoT+A contrastive data points as the training set for SFT Stage 1, thereby injecting the capability to determine when to trigger 
$\langle \text{action} \rangle $. Subsequently, for SFT Stages 2 and 3, we combined 5,000 CoA data points with 1,000 CoT data points to instill the capability of how to execute 
$\langle \text{action} \rangle $. Including 1,000 CoT samples ensures the model retains its ability to solve tasks without external tools, preventing over-reliance on actions and balancing reasoning with tool usage. This mixed strategy ensures that the model can adaptively decide whether to utilize external tools based on task complexity and knowledge requirements, thereby maintaining stable performance across diverse types of questions.

\subsubsection{SFT-stage1: CoT+A}
In this subsection, we describe our training approach for teaching reasoning models to decide when to switch from internal reasoning to external action—a process that can be framed as contrastive learning over sequence patterns. During this learning process, the policy model \(\pi_{\theta}\) distinguishes between two distinctly different sequence output modes: one mode is initiated by the thinking-type token \(\langle \text{think} \rangle\) and ultimately ends with an action-type token \(\langle \text{action} \rangle\), triggering interaction with the external environment; the other mode includes only the model’s internal reasoning behaviors initiated by the \(\langle \text{think} \rangle\) token. Both modes share the same prefix context. 

We denote by \(x_{\text{chosen}} = [x_1, x_2, \dots, x_n]\) the sequences for which we intend to boost the probability under the policy model, and by \(x_{\text{rejected}}\) the sequences whose probabilities we aim to reduce. In most cases, we consider sequences containing the \(\langle \text{action} \rangle\) token as \(x_{\text{chosen}}\). However, to avoid reinforcing model biases and to help the model correctly learn when to act, we also include some sequences containing only the \(\langle \text{think} \rangle\) token and ultimately arriving at the correct answer as part of \(x_{\text{chosen}}\).

Given a context \(\mathbf{c}\), we compute the log logprobability of each target output sequence. After normalization, the result can be calculated as:
$$
P(x\ |\ \mathbf{c}) = \frac{1}{|x|} \sum_{i=1}^{|x|}\text{log}\ \pi_{\theta}(x_i\ |\ \mathbf{c}, x_{<i})
$$

where \(|x|\) denotes the length of the sequence, and \(\pi_{\theta}(\cdot)\) represents the probability output by the current model for the given sequence. The contrastive loss function used at this stage is thus formulated as:
$$
L_{contra} = -\log\,\sigma\left[P(x_{\text{chosen}}\ |\ \mathbf{c}) - P(x_{\text{rejected}}\ |\ \mathbf{c})\right]
$$

where $\sigma$ is the sigmoid function. This loss encourages the model to assign higher probabilities to chosen sequences than to rejected sequences.

During training, we observed a phenomenon: both chosen and rejected sequence probabilities tend to decrease simultaneously, potentially leading to model collapse. To prevent this, we incorporate an auxiliary supervised fine-tuning loss:

$$
L_{aux}=-P(x_{\text{chosen}}\ |\ \mathbf{c})
$$

This auxiliary loss encourages the model to maintain high probabilities for chosen sequences while differentiating them from rejected sequences. The final loss combines these components:

$$L = L_{contra} + \alpha \cdot L_{aux}$$

where $\alpha$ is a coefficient controlling the contribution of the auxiliary loss.

\subsubsection{SFT-stage2: CoT + CoA (w/ observation mask)}
After the previous stage, the policy model $\pi_{\theta}$ has learned to decide at the individual step level when to trigger actions for external assistance; however, it has not yet captured complete decision-making trajectories. Therefore, at this stage,we further train the model via supervised fine-tuning on synthetic complete decision-making trajectories that integrate both CoT and CoA components. This enhances the model's capability to couple multiple reasoning and action steps, teaching it a behavioral paradigm of alternating reasoning with action, while preserving its original reasoning capabilities as much as possible. Additionally, to avoid interference from external feedback during learning, we mask out loss contributions from tokens representing external feedback.

Given the task context \(\mathbf{tc}\) and the complete decision-making trajectory \(x=[x_1, x_2, \dots, x_n]\), where each \(x_i \in \{\langle \text{think} \rangle,\langle \text{action} \rangle, \langle \text{answer} \rangle\}\), the loss function at this stage is computed as follows:
$$
L = \frac{1}{|x|}\sum_{i=1}^{|x|} I_{x_i \neq \mathbf{o}} \left[\ \log\,\pi_{\theta}(x_i\ |\ \mathbf{tc}, x_{<i})\ \right]
$$

Here, \( I_{x_i \neq \mathbf{o}} \) indicates tokens that do not correspond to external feedback, and we only consider the numerical terms associated with these tokens in the loss calculation. 

\subsubsection{SFT-stage3: CoT + CoA (w/o observation mask)}
In this stage, we modify the training process by including the loss for external environment feedback tokens in the complete trajectories. During training, the model is tasked with optimizing both its internal reasoning and action outputs while also accurately predicting external feedback. This joint training effectively serves as implicit learning of an internal world model, thereby offering a more robust initialization for the subsequent reinforcement learning stage.

\subsection{Reinforcement Learning}

The reinforcement learning stage in the AutoCoA framework aims to internalize the Chain-of-Action generation capability into the reasoning model, enhancing its multi-turn tool usage capacity. Building on the Supervised Fine-Tuning stage, RL employs Group Relative Policy Optimization (GRPO) to refine the model's ability to interleave CoT and CoA sequences. The approach utilizes a precise matching rule reward system, ensuring that the model not only reaches the correct answer but also adheres to predefined formatting requirements.

\subsubsection{Reward Function}
The reward function in RL is strictly based on two criteria: exact match with the ground truth and adherence to the required format.  The model receives rewards only when the generated output exactly matches the expected result. Additionally, a format penalty is introduced to enforce the correct structure of the output sequence.

\begin{itemize}
    \item \textbf{Exact matching reward}: A positive reward is awarded only when the final output is an exact match with the ground truth. This approach encourages the model to maintain high precision in generating responses.
    \item \textbf{Format penalty}: A penalty is imposed if the model's output does not adhere to the required format, such as missing critical tags like \( \langle \text{think} \rangle \). This mechanism ensures the model generates well-structured CoT and CoA sequences.
\end{itemize}

\subsubsection{Policy Optimization using GRPO}
The model is optimized using the Group Relative Policy Optimization (GRPO) algorithm, which is adapted to the generative nature of CoT and CoA sequences. GRPO evaluates a group of potential actions at each step and optimizes the policy based on the relative performance of these actions within the group.

For each state, GRPO samples a set of actions and computes the group-relative advantage:

\[
A_t = r_t - \frac{1}{K} \sum_{k=1}^K r_{t,k}
\]

Where \( r_t \) is the reward of the selected action, and \( K \) is the size of the action group. This advantage function guides the model to prefer actions that outperform others in the same context.

The objective function for GRPO is:

\[
J(\theta) = \mathbb{E}_{\tau \sim \pi_\theta} \left[\sum_{t=0}^T \log \pi_\theta(a_t | s_t) A_t \right]
\]

This method ensures that policy updates are driven by the relative quality of actions, promoting stability and efficiency during training. 

\subsubsection{RL-stage1: CoT + CoA (simulated environment)}
In the first RL stage, the model interacts with a \textbf{simulated environment} using its internal world model developed during the SFT phase. This simulated environment acts as a controlled, low-cost, and highly efficient sandbox where the model can freely explore a diverse range of action strategies.

During this stage, the model generates hypothetical tool interactions and simulates the responses using its learned internal world model. GRPO is utilized to explore a wide array of potential CoA strategies through group-based sampling and rule-based rewards. Through iterative refinement, the model learns to balance reasoning and action, improving its ability to decide when and how to invoke external tools.

\subsubsection{RL-stage2: CoT + CoA (real environment)}
In the second RL stage, the agent model transitions to the \textbf{real environment}, where it interacts with actual tools to obtain real-world feedback. This phase focuses on adapting the strategies initially learned from the simulated environment to dynamic and unpredictable real-world scenarios. 

Compared to the previous stage, in this stage, the model executes real tool invocations (e.g., web searches) and replaces the simulated responses with actual feedback obtained from these tools. This staged approach enables the agent model to maintain its internalized CoA generation capability, ensuring strong performance in both controlled and open-ended task environments.

\section{Experiments}
\subsection{Settings}

The evaluations in this experiment encompass a range of open-domain QA tasks, including 2 single-hop QA datasets and 3 multi-hop QA datasets.

\textbf{Single-hop QA datasets:}
\vspace{-2mm}
\begin{itemize}
    \item  \emph{Natural Questions (NQ)}~\cite{kwiatkowski2019natural} consists of real-world Google search queries with answers sourced from Wikipedia articles, providing a benchmark for answering straightforward questions. 
    \item \emph{TriviaQA}~\cite{joshi2017triviaqa} is a large-scale dataset derived from trivia websites and competitions, characterized by complex entity relationships and challenging question formulations. 
\end{itemize}

\textbf{Multi-hop QA datasets:} 
\vspace{-2mm}
\begin{itemize}
\item \emph{HotpotQA}~\cite{yang2018hotpotqa} is the pioneering large-scale dataset that requires reasoning across multiple Wikipedia paragraphs to derive answers.
    \item  \emph{2WikiMultihopQA (2WIKI)}~\cite{ho2020constructing} enhances multi-hop question answering by offering explicit reasoning paths. 
    \item \emph{MuSiQue}~\cite{trivedi2022musique} presents 2-4 hop questions constructed by combining elements from five existing single-hop datasets, testing the ability to integrate multiple information sources.
    \item \emph{Bamboogle}~\cite{press2022measuring} introduces a set of complex questions that major search engines, including Google, often answer incorrectly, offering a robust evaluation of models' compositional reasoning across diverse domains.
\end{itemize}

In this experiment, we use two evaluation metrics:
\vspace{-2mm}

\begin{itemize}
    \item \textbf{Exact Match accuracy (EM)} : Counts as correct only when the model's prediction is exactly the same as the ground truth. This metric strictly evaluates the model's precision.
    
    \item \textbf{LLM-based accuracy (LLM)}: Uses LLM (Qwen2.5-14B-instruct) to determine if the model's prediction semantically matches the ground truth, allowing for more diverse correct answers.
\end{itemize}

We selected a small-scale reasoning model, R1-Distill-Qwen-7B, for AutoCoA training. For parameter optimization, we utilized LLaMA-Factory~\cite{zheng2024llamafactory} as the framework for Supervised Fine-Tuning and verl~\cite{sheng2024hybridflow} as the framework for Reinforcement Learning. All RL training was conducted on a single node with Nvidia H20 GPUs, utilizing sequence parallelism.

\subsection{Compared methods}

We conducted a comprehensive evaluation of our AutoCoA framework by comparing it with various baseline methods across different categories. This section provides a brief introduction to these methods.

\textbf{Model only (w/o action)}
\vspace{-1mm}

To establish a baseline for the inherent knowledge capacity of foundation models without retrieval augmentation, we evaluate several reasoning-focused models in a zero-shot setting:
\vspace{-2mm}
\begin{itemize}
    \item \emph{R1-Distill-Qwen-7B} : the initial policy model used for AutoCoA training.
    \item \emph{Deepseek-R1} : The original 670-B large reasoning model with enhanced reasoning capabilities, representing state-of-the-art performance without retrieval augmentation.
\end{itemize}

\textbf{Agent workflow}

We implemented traditional agent workflows (ReAct) on different types of base models to evaluate the effectiveness of action frameworks across model categories:
\vspace{-1mm}

\begin{itemize}
    \item \emph{Qwen-7B-Base(ReAct)} : A base language model configured with ReAct prompting techniques to generate reasoning and action sequences. This represents a standard baseline for agentic workflows without specialized reasoning capabilities.
    \item \emph{R1-Distill-Qwen-7B(ReAct)} : A reasoning-enhanced model configured with ReAct prompting. This approach combines enhanced reasoning capabilities with a
structured action framework.
\end{itemize}
    Both approaches follow standard ReAct protocols where external tool calls are
explicitly structured through prompting rather than being internalized as model
capabilities.

\textbf{Agent model}

We evaluated several variants of our agent model to understand the impact of each training stage:
\vspace{-1mm}
\begin{itemize}
    \item \emph{SFT-stage1} : The model after completing only the CoT+A stage, which focuses on teaching when to $\langle action \rangle$ using contrastive learning at the step level. This stage used 1.5K CoT+A contrastive data points.
    \item \emph{SFT-stage2} : The model after completing only the CoT+CoA(w/ observation mask) stage, which focuses on trajectory-level action patterns without observation mask prediction. This stage used a mix of 5K CoA and 1K CoT data points.
    \item  \emph{SFT-stage1\&2} : A combined approach where the model undergoes both SFT-Stage 1 and SFT-Stage 2 sequentially, benefiting from both step-level action triggering and trajectory-level pattern learning.
    \item  \emph{SFT-stage1\&2\&3} : The complete SFT pipeline, including the CoT+CoA(w/o observation mask) stage that enables the model to predict environment responses alongside actions, effectively learning an implicit world model. This stage used the same data mix as SFT-stage2.
\end{itemize}

We compared the performance of different RL settings and methods. All training was conducted with 96 optimization steps on the same 4,608 problems:
\vspace{-1mm}

\begin{itemize}
    \item \emph{RL-stage2}: In this setting, we utilize in-context learning to guide a model without SFT through CoA-based reasoning. An additional 50\% optimization steps were added for this training configuration.
    \item \emph{SFT-stage1\&2\&3 + RL-stage1}: The model is trained exclusively in the simulated environment, using \emph{SFT-stage1\&2\&3} as the initial policy model.
      \item \emph{SFT-stage1\&2\&3 + RL-stage1\&2}: The model is first trained for the initial 80 steps (5/6) in the simulated environment, then switched to the real environment for the rest 1/6. The initial policy model is set to \emph{SFT-stage1\&2\&3}.
    \item \emph{SFT-stage1\&2 +RL-stage2}: The model is trained exclusively in the real environment, using \emph{SFT-stage1\&2} as the initial policy model.
    
\end{itemize}

\subsection{Results}
The main results are summarized in Table~\ref{tab:results}. Observations include: 
\begin{itemize}
    \item \emph{Agentic workflow vs. agent model}. After CoA learning, all AutoCoA variants substantially outperform the initial policy model with the ReAct workflow. This validates the effectiveness of end-to-end optimization for task-oriented  when and how to act during reasoning.
    \item \emph{Agent model: SFT-stage1\&2 vs. SFT-stage2}. Separating CoA fine-tuning into \emph{when-to-act} (stage1) and \emph{how-to-act} (stage2) sub-stages enhances the model's ability to learn CoA patterns, enabling better transitions between thought and action.
    \item \emph{Agent model: RL-stage2}: Compared to the initial policy, direct RL shows some improvement but is constrained by the initial policy’s baseline capacity, resulting in marginal gains. 
Direct RL application shows marginal improvements over the initial policy, constrained by the policy's baseline capacity.
\item \emph{Agent model: SFT-stage1\&2\&3 + RL-stage1 vs. SFT-stage1\&2\&3 +RL-stage1\&2}. After extensive (5/6) simulated environment training (RL-stage1), a small portion (1/6) of real-world interaction (RL-stage2) improves adaptation to dynamic environments, further boosting accuracy.
\item \emph{Agent model: SFT-stage1\&2 + RL-stage2 vs. SFT-stage1\&2\&3 +RL-stage1\&2}. Compared to full real-world interaction, initial simulated environment training sacrifices some performance. However, it significantly reduces interaction costs. Achieving acceptable results with only 1/6 real interactions demonstrates the feasibility of internal world modeling. Future work is needed to improve world model fidelity and leverage the saved costs for advanced training paradigms.
\end{itemize}

\begin{landscape}
\begin{table}[ht]
\centering
\parbox{1.45\textwidth}{\caption{Performance comparison between different QA methods. \textbf{Bold} indicates the highest value, while \underline{underline} indicates the highest value among the different settings of \emph{AutoCoA}.}\label{tab:results}\vspace{1.5mm}}

    \begin{tabular}{lcccccccccccccc}
    \toprule
    \multirow{4}{*}{Method} & \multicolumn{4}{c}{Single-hop QA} & \multicolumn{8}{c}{Multi-hop QA} & \multicolumn{2}{c}{Average} \\
    \cmidrule(lr){2-5} \cmidrule(lr){6-13} \cmidrule(lr){14-15}
    & \multicolumn{2}{c}{NQ} & \multicolumn{2}{c}{TriviaQA} & \multicolumn{2}{c}{HotpotQA} & \multicolumn{2}{c}{2WIKI} & \multicolumn{2}{c}{MuSiQue} & \multicolumn{2}{c}{Bamboogle} & \multicolumn{2}{c}{} \\
    \cmidrule(lr){2-3} \cmidrule(lr){4-5} \cmidrule(lr){6-7} \cmidrule(lr){8-9} \cmidrule(lr){10-11} \cmidrule(lr){12-13}
    & EM & LLM & EM & LLM & EM & LLM & EM & LLM & EM & LLM & EM & LLM & EM & LLM \\
    \midrule
    \multicolumn{15}{l}{\textit{\textbf{Model only (w/o action)}}} \\ [2pt]
    R1-Distill-Qwen-7B      & 5.9 & 9.8 & 17.3 & 21.1 & 11.7 & 14.1 & 23.3 & 25.9 & 1.1 & 2.5 & 11.7 & 13.1 & 11.8 & 14.4 \\
    Deepseek-R1\textcolor{blue}{*}          & \textbf{39.0} & \textbf{54.2} & \textbf{78.2} & \textbf{85.6} & \textbf{44.8} & \textbf{53.0} & \textbf{52.4} & \textbf{55.8} & \textbf{21.2} & \textbf{27.4} & \textbf{56.8} & \textbf{63.2} & \textbf{49.5} & \textbf{60.3} \\
    \midrule
    \multicolumn{15}{l}{\textit{\textbf{Agentic workflow}}} \\[2pt]
    Qwen-7B-Base (\emph{ReAct})  & 18.0 & 37.8 & 35.2 & 48.8 & 17.2 & 28.4 & 24.0 & 30.8 & 4.2 & 8.4 & 19.2 & 24.8 & 19.6 & 29.8 \\
    R1-Distill-Qwen-7B (\emph{ReAct})  & 15.1 & 20.9 & 25.7 & 30.1 & 15.1 & 18.6 & 21.3 & 22.9 & 1.9 & 3.9 & 12.3 & 14.7 & 15.2 & 18.5 \\
    \midrule
    \multicolumn{15}{l}{\textit{\textbf{Agent model: SFT}}} \\[2pt]
    SFT-stage1 & 22.7 & 32.2 & 40.0 & 47.8 & 21.2 & 28.3 & 26.4 & 29.6 & 4.8 & 8.2 & 20.2 & 22.8 & 22.7 & 29.1 \\
    SFT-stage2 & 25.0 & 33.1 & 42.6 & 48.2 & 31.7 & 39.2 & 43.8 & 48.4 & 8.5 & 12.1 & 24.3 & 27.2 & 29.5 & 35.6 \\
    SFT-stage1\&2 & 25.0 & 34.3 & 45.6 & 51.3 & 33.0 & 40.8 & 48.1 & 52.4 & 9.0 & 12.4 & \underline{30.4} & \underline{33.3} & 32.0 & 38.5 \\
    SFT-stage1\&2\&3 & 23.8 & 33.1 & 44.7 & 51.2 & 30.1 & 36.5 & 49.2 & 52.0 & 10.5 & 14.7 & 28.8 & 32.0 & 31.3 & 37.7 \\
    \midrule
    \multicolumn{15}{l}{\textit{\textbf{Agent model: RL}}} \\[2pt]
    RL-stage2 & 22.8 & 30.0 & 37.5 & 43.2 & 23.2 & 28.2 & 33.2 & 33.8 & 5.6 & 8.2 & 20.0 & 23.2 & 23.7 & 27.8 \\
    SFT-stage1\&2\&3+RL-stage1 & 26.0 & 33.4 & 47.2 & 53.8 & 31.4 & 40.0 & 48.6 & 51.4 & 9.6 & 12.4 & 27.2 & 32.0 & 31.7 & 37.2 \\
    SFT-stage1\&2\&3+RL-stage1\&2 & \underline{28.6} & \underline{37.0} & 48.6 & 54.0 & 33.2 & 41.2 & \underline{49.6} & \underline{53.6} & \underline{12.6} & \underline{16.2} & 28.0 & 31.2 & 33.4 & \underline{38.9} \\
    SFT-stage1\&2+RL-stage2 & 28.4 & 34.6 & \underline{52.4} & \underline{56.6} & \underline{35.6} & \underline{42.8} & 48.4 & 51.2 & 12.0 & 15.2 & 26.4 & 30.4 & \underline{33.9} & 38.5 \\
    
    \bottomrule
    \end{tabular}
    
    \parbox{1.45\textwidth}{\vspace{1.5mm}
    \textcolor{blue}{*} \textit{The results reflect R1's knowledge memorization rather than action capabilities, due to its large parameter size and the outdated knowledge in QA questions. On new questions, R1's accuracy is very low. See Appendix.A for details.} 
    }
\end{table}
\end{landscape}

\subsubsection{Improving long-horizon task execution}
To evaluate the role of CoA learning in tackling tasks that require long-horizon reasoning and action, we analyzed model performance before and after CoA learning as the number of executed actions ($\#action$) increased.

\begin{figure}
\centering
\subfigure[Ratio of correctly completed tasks.]{
\begin{minipage}[t]{0.8\linewidth}
\centering
\includegraphics[width=\linewidth]{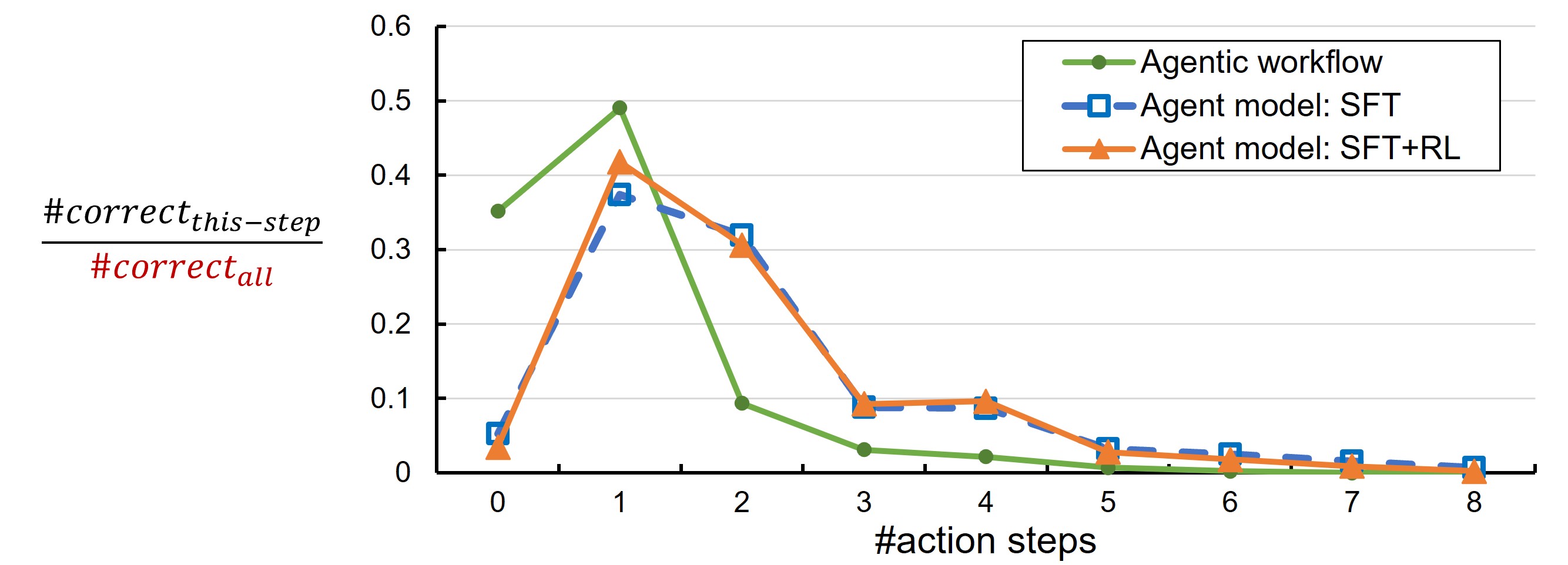}
\end{minipage}%
}%
\hfill
\subfigure[Task success rate.]{
\begin{minipage}[t]{0.8\linewidth}
\centering
\includegraphics[width=\linewidth]{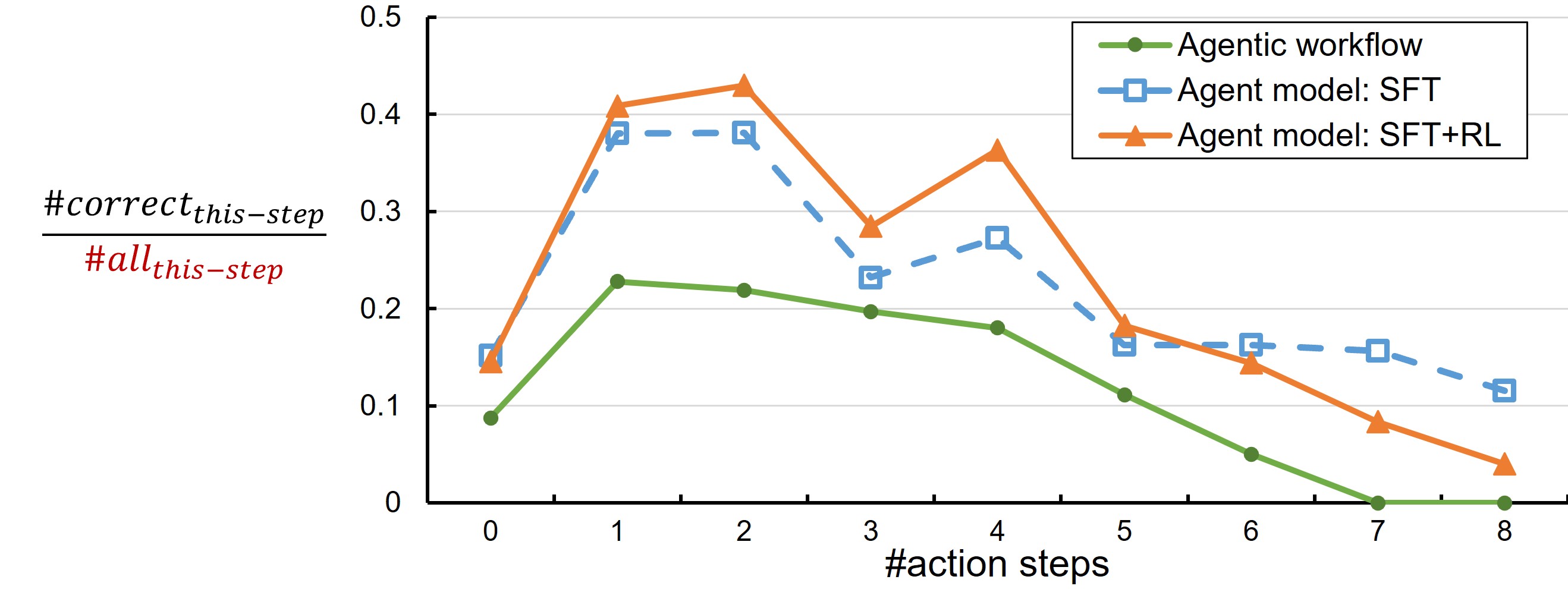}
\end{minipage}%
}%
\centering
\caption{Analysis of hong-horizon execution capabilities.}
\label{fig:3}
\end{figure}

Fig.~\ref{fig:3}(a) illustrates the distribution of correctly completed tasks across different $\#action$. Without CoA learning, the initial policy model primarily succeeded on tasks requiring only 1 or 2 actions. After CoA learning, agent models demonstrated the ability to support longer thought/action rounds, enabling them to complete more complex tasks.

We further examined task success rates by $\#action$ (Fig.~\ref{fig:3}(b)). The results reveal a consistent trend with Fig.~\ref{fig:3}(a): Initial policy with workflow shows declining success rates as $\#action$ increases, whereas agent models maintain a relatively high accuracy at 5 actions. This demonstrates that task-oriented end-to-end training enables agent models to learn improved interleaving patterns between thought and action, with strong consistency in long-context scenarios as the number of actions increases.

\section{Conclusion and Future Work}
We propose AutoCoA, an agent model learning approach to enhance the multi-turn tool use capability on top of reasoning models. Through end-to-end fine-tuning, the derived agent model demonstrates a superior ability to determine when and how to act in pursuit of the ultimate goal, outperforming traditional agentic workflows in goal-oriented decision-making.

Methodologically, this report represents an initial exploration into learning CoA (Chain-of-Action). It serves as a preliminary attempt, with a wide range of potential approaches yet to be explored, including pure reinforcement learning (RL), improved action-inclusive sampling, and loss design for trajectories that interleave CoT (Chain-of-Thought) and CoA.

From a task perspective, the next steps are extending the application scope: (1) Agent model for open-ended tasks, exploring methods for implementing agent models capable of handling open-ended tasks with rubrics for grading; (2) AgentRFT (Reinforcement Fine-Tuning), building on our prior work with OpenRFT~\cite{zhang2024openrft} to tailor the agent foundation model to meet specific task objectives and leverage specialized tools within particular scenarios. 

\subsection{Agent Roadmap}
OpenAI has long pursued the vision of seamless environmental interaction. \emph{Operator} stems from its early key project ``Computer Use'', and \emph{Deep Research} continues its goal since ``Plugins'' and ``GPTs''. However, the earlier attempts were constrained by the limited capabilities of models at the time. 

Today, with advancements in language modeling and logical reasoning, Deep Research has tackled highly complex tasks using only basic tools like web search and document parsing. This demonstrates a universal recipe for agent development: once a model's reasoning capacity surpasses a certain threshold, providing it with a few simple tools and a verifiable task objective within a controlled environment enables what the Claude 3.7 official blog describes as ``action scaling''. The fact that OpenAI has began to assess model capabilities via replaceable human expert hours is particularly noteworthy. Even if a model only possesses an average level of human intelligence and tool usage ability, combined with the efficiency of AI in information acquisition and processing, the enhancement to productivity would be explosive.

\begin{figure}[t] 
  \centering 
  \includegraphics[width=0.8\textwidth]{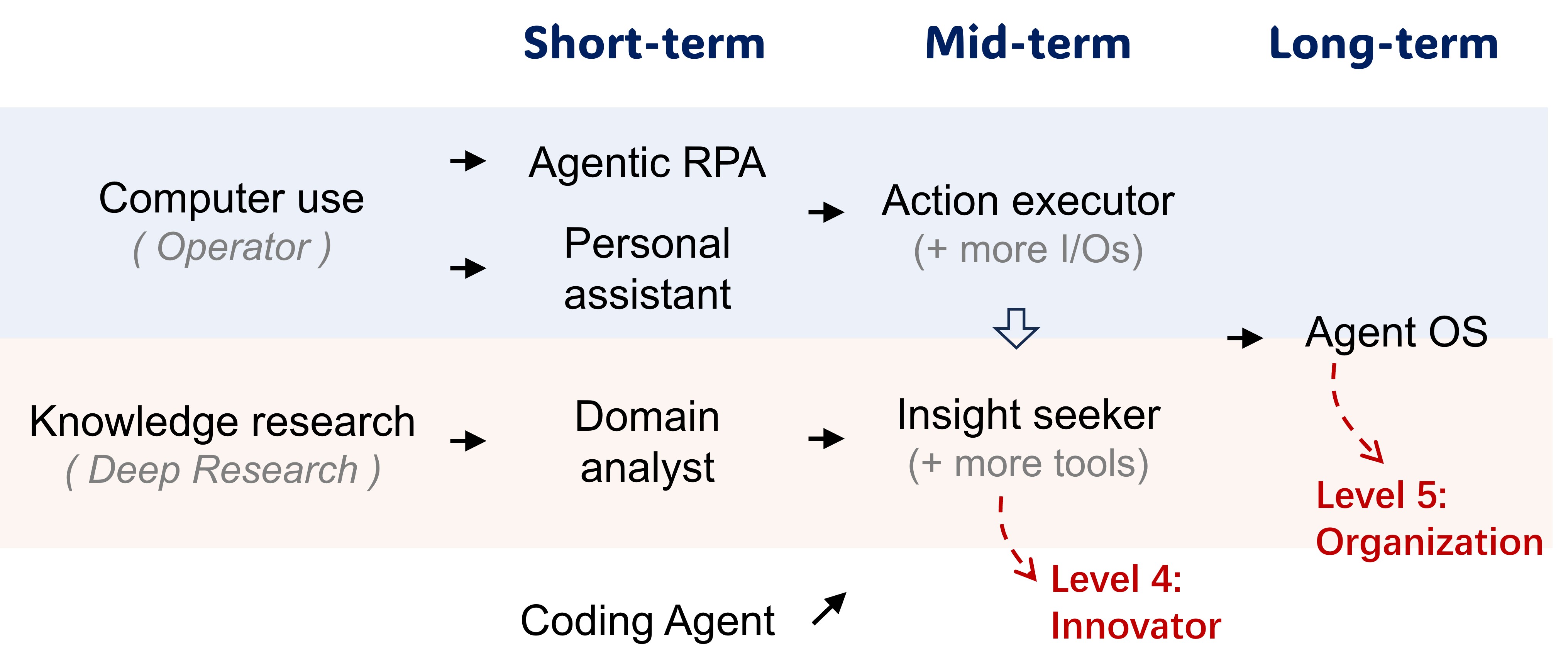} 
  \caption{Agent roadmap.} 
  \label{fig:x} 
\end{figure}

Operator and Deep Research exemplify OpenAI’s agent paradigm—combining execution (Operator as the ``hands'') and reasoning (Deep Research as the ``brain'') to achieve dynamic programming, autonomous tool usage, asynchronous processing, and automatic task completion. We categorize their functionalities as \emph{Computer Use} and \emph{Knowledge Research}, respectively, and envision a three-phase roadmap for agent development (Fig.~\ref{fig:x}).

\subsubsection{Short-term: Applications}
By enhancing reasoning foundation models with basic tool-usage capabilities to create agent foundation models, practical applications are enabled through methods like Reinforcement Fine-Tuning (RFT). Below are the key scenario applications for this stage:

\textbf{Agentic RPA}

\vspace{-1.5mm}
A natural evolution of traditional Robotic Process Automation (RPA), Agentic RPA integrates learning and reasoning to automate enterprise workflows (e.g., ERP/CRM systems). Unlike rule-based RPA, which struggles with unstructured data or dynamic environments, Agentic RPA adapts to evolving business needs, can tackle more complex and flexible tasks.

\textbf{Personal assistant}

\vspace{-1.5mm}
Focused on mobile-centric, personalized interactions, this agent leverages behavioral logs and personal knowledge bases to act as a user’s ``digital twin.'' Future extensions include proactive behavior (predicting needs rather than passive responses) and multimodal context awareness (perceiving and interpreting environmental cues to refine task execution).

\textbf{Domain analyst}

\vspace{-1.5mm}
Deep Research currently faces two main limitations: restricted open-domain data access and shallow expert analysis. For specialized fields such as law, medicine, and research, addressing these gaps requires fine-tuning on proprietary data and domain knowledge. This may enable the leap into actionable insights, supporting real-world analytical consulting and decision-making.

\textbf{Coding agent}

\vspace{-1.5mm}
In addition to applying computer use and knowledge research capabilities, another critical short-term direction is automating software engineering, as exemplified by projects like \emph{Devin} and \emph{Manus}. Coding agents treat programming not as an end goal but as a tool-use mechanism (e.g., automating API integrations). Whether deployed internally or as standalone products, coding agents represent foundational infrastructure for scaling agent capabilities.

\subsubsection{Mid-term: Expanding Capabilities}
Agent evolution in the mid-term will focus on two parallel directions: (1) Action executor, enhancing task execution by accessing a broader range of I/O interfaces, evolving into ``more capable hands and feet.'' (2) Insight seeker, deepening information processing through expanded tool integration, becoming a ``more comprehensive brain.''

\textbf{Action executor}

\vspace{-1.5mm}
Beyond integrating more software application APIs, agent task execution will extend from software operations to the physical world, requiring enhanced interfaces for both input perception and output execution. IoT device interfaces are typical inputs. By connecting to IoT devices in smart homes and factories, Agent can better sense the environment, and understand human behavior and needs for automated task execution. Output execution interfaces will involve remote operation and control for physical-world automation.

\textbf{Insight seeker}

\vspace{-1.5mm}
At the Deep Research release, OpenAI noted that search is just the first step, and hopes to integrate more tools in the future to let the Agent ``autonomously discover new knowledge''. Possible tools include information acquisition (data sources, unstructured information parsing) and information processing (general visualization, database analysis, and domain-specific tools). 

Currently, knowledge research agents mainly focus on discovery and information integration within the language space. Innovation and insight require more input. The current input is provided by human feedback in the human-machine collaboration paradigm. In the future, inputs could come from new findings through tool experiments or feedback from multi-modal intelligence.

At that time, information processing will no longer start with explicit user requests nor take the form of simple Q\&A. Instead, the role will shift: we will provide agents with data, experimental materials, and simulation environments, allowing them to autonomously explore, analyze, deduce, and generate new insights. With the continued development of action executors and coding agents, the insight seeker could further evolve toward the fourth stage: \textbf{Innovator}.

\subsubsection{Long-term: Agent OS}
Web pages and apps have served as the application carriers for internet and mobile internet technologies, respectively, while agents are now emerging as the carriers for artificial intelligence technologies. From a computational perspective, agent models function as the CPU in traditional computer architecture, handling core computation and task orchestration. Tools function as software programs, executing specific operations, and short-term and long-term memory correspond to RAM and hard drives. Together, these components form the basis of what could be termed an \emph{agent OS}.

However, building an agent OS presents significant challenges. Much like the von Neumann architecture relies on the unified storage of programs and data alongside a complete instruction set, agent OS requires unified representations for multimodal I/O, tools, and memory, as well as an instruction set tailored for agents. This poses difficulties for current Transformer models, which depend on token-based sequential representations, and for natural language-based instruction systems.

Furthermore, data representation and organizational structures need to be adapted to suit agent models. GUIs like web pages and apps, designed for human interaction, are not the most efficient for agents. Agent protocol is required to define how agents interact with tools, vector databases, and other agents, enabling operations at a more foundational level to fully harness their strengths. 

As the agent ecosystem matures, multi-agent collaboration will become pivotal. Agents will dynamically coordinate in complex projects--sharing intermediate results, delegating subtasks, and self-organizing like a human team. This evolution from individual competence to collective intelligence, marked by enhanced autonomy and cooperative problem-solving, may signal the emergence of the fifth stage: \textbf{Organization}. 


\bibliography{iclr2025_conference}
\bibliographystyle{iclr2025_conference}

\newpage
\appendix
\section{Appendix}

\renewcommand{\arraystretch}{1.5}

\begin{table}[h]
\centering
\caption{Performance Comparison across Temporal Knowledge Domains}
\label{tab:dataset_comparison}
\small
\renewcommand{\arraystretch}{1.2}
\begin{tabular}{p{2.5cm}cccccccc} 
\toprule
\multirow{2}{*}{Dataset} & \multicolumn{4}{c}{Recent (\emph{Unknown})} & \multicolumn{4}{c}{Popular\& Old (\emph{Known})} \\
\cmidrule(lr){2-5} \cmidrule(lr){6-9}
& \multicolumn{2}{c}{Main Question} & \multicolumn{2}{c}{Sub-question} & \multicolumn{2}{c}{Main Question} & \multicolumn{2}{c}{Sub-question} \\
\cmidrule(lr){2-3} \cmidrule(lr){4-5} \cmidrule(lr){6-7} \cmidrule(lr){8-9}
Model & EM & LLM & EM & LLM & EM & LLM & EM & LLM \\
\midrule
DeepSeek-R1-Distill-Qwen-7B & 0.026 & 0.052 & 0.017 & 0.039 & 0.052 & 0.073 & 0.231 & 0.267 \\[2pt]
DeepSeek-R1 & 0.078 & 0.098 & 0.079 & 0.119 & 0.615 & 0.646 & 0.899 & 0.944 \\
\bottomrule
\end{tabular}
\end{table}


\subsection{Datasets}

This work evaluates how model knowledge memorization impacts multi-hop reasoning accuracy. The experiments are structured as follows:

1. \textbf{Unknown knowledge evaluation}:  
   A test set of 192 three-hop questions (576 sub-questions) was constructed using time-sensitive corpora sourced from 97 sports-related reports published by CCTV News between January 16, 2025, and March 7, 2025. This dataset ensures that the models were not exposed to the data during training, allowing for an evaluation of their ability to handle unseen temporal knowledge.

2. \textbf{Known knowledge evaluation}:  
   For comparison, we used the MINTQA-POP benchmark dataset, proposed by \cite{he2024mintqa}. We use a subset of 192 three-hop popular questions (576 sub-questions) was selected through stratified sampling. This dataset covers high-frequency knowledge domains familiar to the models from wiki datasets, enabling an assessment of model performance on known knowledge.

\subsection{Experimental Results}

The results highlight the models' distinct performance across the two evaluation scenarios:

1. \textbf{Performance on questions requiring known knowledge}:  
   R1 achieved \emph{61.5\%} (EM) and \emph{64.6\%} (LLM) on MINTQA-POP, vastly outperforming the distilled 7B model's \emph{5.2\%} (EM) and \emph{7.3\%} (LLM). This demonstrates R1's reliance on memorized patterns rather than generalized reasoning--it excels on tasks involving familiar knowledge domains.

2. \textbf{Performance on questions requiring unknown knowledge}:  
   Both models collapsed on unseen temporal data: R1 dropped to  \emph{7.8\%} (EM) and \emph{9.8\%} (LLM), and R1-Distill-Qwen-7B dropped to \emph{2.6\%} and \emph{5.2\%} (LLM). The drastic performance drop highlights a fundamental knowledge cutoff dependence. As noted in Section 4.3, R1's superior performance stems from memorizing outdated wiki facts--when tested on temporal questions requiring very recent knowledge, its accuracy declines significantly, underscoring the necessity of augmenting internal knowledge through tool usage from external environments.

\end{document}